\documentclass[letterpaper, 10 pt, conference]{ieeeconf}  

\IEEEoverridecommandlockouts                              

\usepackage{graphics} 
\usepackage{amsmath} 
\usepackage{cite}
\usepackage{subfigure}
\usepackage{tabulary}
\usepackage{multirow}
\usepackage{multicol}

\usepackage{todonotes}
\graphicspath{{figs/}}
\usepackage[colorlinks,linkcolor=blue,anchorcolor=blue,citecolor=red]{hyperref} 
\usepackage{wrapfig,lipsum,booktabs}
\usepackage{xcolor}
\usepackage{color}
\usepackage{cite}
\usepackage{algorithmic} 
\usepackage[ruled,vlined,commentsnumbered]{algorithm2e}

\usepackage[flushleft]{threeparttable}
\usepackage{gensymb}  
\usepackage{amsmath}
\usepackage{amssymb}

\usepackage{flushend}

\graphicspath{{figs/}}

\usepackage{siunitx}
\usepackage{url}   

\title{\LARGE \bf
Direct Monocular Odometry Using Points and Lines
}

\author{Shichao Yang, Sebastian Scherer 
\thanks{The Robotics Institute, Carnegie Mellon University, 5000 Forbes Ave, Pittsburgh, PA 15213, USA.
		{\tt\small \{shichaoy, basti\}@andrew.cmu.edu}}
}

\begin{document}
\maketitle
\thispagestyle{empty}
\pagestyle{empty}

\begin{abstract}
Most visual odometry algorithm for a monocular camera focuses on points, either by feature matching, or direct alignment of pixel intensity, while ignoring a common but important geometry entity: edges. In this paper, we propose an odometry algorithm that combines points and edges to benefit from the advantages of both direct and feature based methods. It works better in texture-less environments and is also more robust to lighting changes and fast motion by increasing the convergence basin. We maintain a depth map for the keyframe then in the tracking part, the camera pose is recovered by minimizing both the photometric error and geometric error to the matched edge in a probabilistic framework. In the mapping part, edge is used to speed up and increase stereo matching accuracy. On various public datasets, our algorithm achieves better or comparable performance than state-of-the-art monocular odometry methods. In some challenging texture-less environments, our algorithm reduces the state estimation error over 50\%.
\end{abstract}

\section{Introduction}
Visual odometry (VO) and Simultaneous localization and mapping (SLAM) have become popular 
topics in recent years due to their wide application in robot navigation, 3D reconstruction, 
and virtual reality. Different sensors can be used such as RGB-D cameras \cite{huang2011visual},
stereo cameras \cite{Geiger11} and lasers, which could provide depth information for each frame, making it easier 
for state estimation and mapping. However for some applications such as weight constrained micro aerial vehicles \cite{fang2017robust}, monocular cameras are more widely used due to their small size and low cost. Therefore, in this work,
we are aiming at the more challenging monocular VO.

There are typically two categories of VO and vSLAM approaches: (1) feature based methods such
as PTAM \cite{klein2007parallel} and ORB SLAM \cite{mur2015orb}. They rely on feature point extraction 
and matching to create sparse 3D map used for pose estimation by minimizing re-projection 
geometric error. (2) Recently, direct method \cite{engel2013semi}\cite{engel2014lsd} 
also becomes popular. It directly operates on the raw pixel intensity by minimizing photometric error 
without feature extraction. These two methods both have their advantages. Reprojection geometric error 
of keypoints is typically more robust to image noise and large geometric distortions and movement.
Direct method on the other hand, exploits much more image information and can create dense or semi-dense maps.

\begin{figure}[t]
  \centering
   \includegraphics[scale=0.33]{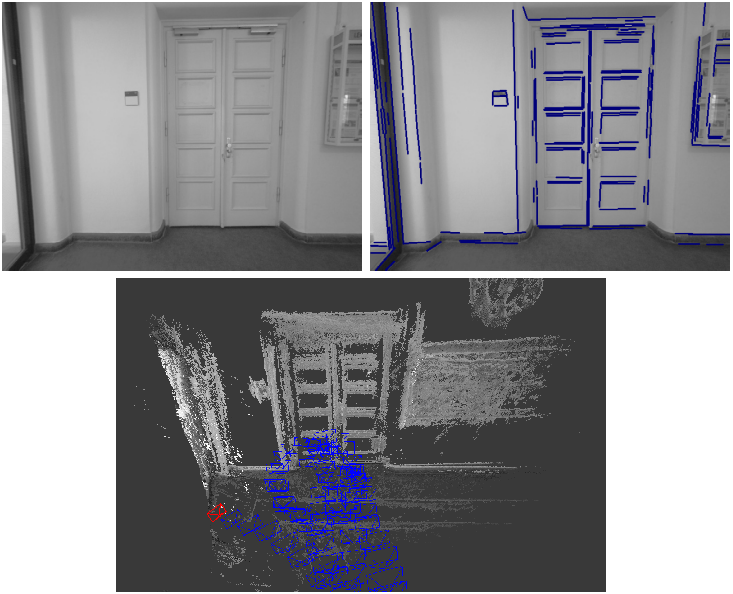}
   \caption{Tracking and 3D reconstruction on TUM mono dataset using our edge based visual odometry. The top image shows the homogeneous wall surface with low image gradients, which is challenging for VO only minimizing photometric error. However, edge shown in blue can still be detected to improve the tracking and mapping performance.}
   \label{fig:method overview}
\end{figure}

In this paper, we utilize points and edges to combine the advantages of the above two approaches.
Edge is another important feature apart from points. It has been used for stereo \cite{gomez2016robust} and RGB-D VO 
\cite{kuse2016robust}, but receives less attention in monocular VO. The detection of edges is less sensitive to 
lighting changes by nature. For example, in a homogeneous environment of Fig. \ref{fig:method overview}, direct method using points only may not work robustly due to small image gradient, but we can still detect many edges shown in blue in the figure which could be used for state estimation and mapping. In our system, we maintain a semi-depth map for the keyframe's high 
gradient pixels as in many direct VO methods \cite{engel2013semi}. We also detect and match edges for 
each frame. Then in the tracking part, we jointly optimize both photometric error and geometric error to the 
corresponding edge if it has. In the mapping part, edges could also be used to guide and speed up the stereo search and also improve depth map quality by edge regularizing. By doing this, the proposed VO can increase the accuracy of state estimation and also create a good semi-dense map. We demonstrate this through various experiments.

In summary, our main contributions are:
\begin{itemize} 
    \item  A real-time monocular visual odometry algorithm incorporating points and edges,
    		   especially suitable for texture-less environments.
    \item  Provide an uncertainty analysis and probabilistic fusion of points and lines in tracking and mapping.
    \item  Develop analytical edge based regularization
	\item  Outperform or comparable to existing direct VO in many datasets.
\end{itemize}

In the following section, we discuss related work. In Section \ref{sec:Problem Description}, we provide the problem
formulation. Tracking and mapping using points and edges are presented in Section \ref{sec:tracking} and
Section \ref{sec:mapping} respectively, which also include probabilistic uncertainty analysis of different observation model. 
In Section \ref{sec:experiments}, we provide experimental comparison with the state-of-art algorithm. Finally, conclusion and future work is discussed in Section \ref{sec:conclusions}


\section{Related Work}
\label{sec:related}
Our algorithm utilizes edges to combine feature based and direct VO. We briefly introduce these three aspects.

\subsection{Feature based VO}
There have been many feature point based VO and SLAM, for example LibVISO \cite{Geiger11} 
and ORB SLAM \cite{mur2015orb}. They first extract image features then track or match them
across images. The camera pose is estimated by solving the PnP (Perspective N-Point Projection) 
problem to minimize geometric error which is more robust to image noise and has a large convergence basin  \cite{mur2015orb} \cite{engel2016direct}. The drawback is that the created map is usually sparse. A separate direct mapping algorithm is required to get a semi-dense map \cite{mur2015probabilistic}.

\subsection{Direct VO}
In recent years, direct method \cite{newcombe2011dtam} also becomes popular. 
It optimizes the geometry directly on the image intensities without any feature extraction so 
it can work in some texture-less environments with few keypoints. It has been used for real-time 
application of different sensors for example DVO for RGB-D cameras \cite{kerl2013dense} and 
LSD SLAM for monocular cameras \cite{engel2014lsd}. The core idea is to maintain a
semi-dense map for keyframes then minimize the photometric error which is a highly non-convex function thus
it requires good initial guess for the optimization. In between direct and feature based methods, 
SVO combines direct alignment and feature points and can be used for high frame rate cameras.

\subsection{Edge based VO}
Edges are another important feature apart from points especially in man-made environments. Edges
are more robust to lighting changes and preserve more information compared to single points.
Line-based bundle adjustment has been used in SLAM or SfM \cite{eade2009edge} \cite{klein2008improving}
which are computationally expensive and require at least three frames for effective optimization.
Line-based VO without bundle adjustment has recently been used for stereo cameras \cite{witt2013robust}
\cite{gomez2016robust} and RGB-D \cite{lu2015robust} \cite{kuse2016robust} and monocular cameras
\cite{jose2015realtime}\cite{gomez2016pl}. Kuse \textit{et al.} minimize the geometric error
to its nearby edges pixels through distance transform \cite{kuse2016robust}  which might cause wrong matching due to false detected
edges and broken edges while our line segment matching could greatly reduce the error. 
Some works only minimize geometric error of two edge endpoints \cite{gomez2016robust} \cite{lu2015robust} 
which may generate large error for monocular cameras due to inaccurate depth estimation.

\section{Problem Description}
\label{sec:Problem Description}
\subsection{System Overview}
Our algorithm is a frame to keyframe monocular VO. We maintain a semi-depth map for the high gradient pixels in the keyframe. Then for each incoming new frame, there are three steps. First, detect line segments and match them with the keyframe's edges. The second step is camera pose tracking. We minimize a combination of pixel photometric error and geometric reprojection error if the pixel belongs to an edge. Lastly, we update the depth map through variable baseline stereo. Edges are used to speed up the stereo search for those edge pixels and also improve reconstruction through an efficient 3D line regularization.

\if(0)
\begin{figure*} 
\begin{center}
  \scalebox{.4}{\includegraphics{track_iter2.png}}
\end{center}
\caption{Tracking iterations for two images in \textit{fr3/cabinet\_big} dataset. The top two images are the reference frame and current frame which are 41 frames apart (about 1.4s). The bottom four images shows the tracking process as optimization iterations goes. We can see that the reprojected edge pixels shown in color gradually coincides with the true edges in gray. Different color represents belongs to
different edges.}
\label{fig:track iters}
\end{figure*}
\fi

\begin{figure*}
  \centering     
  \subfigure[]
    { \includegraphics[width=45mm]{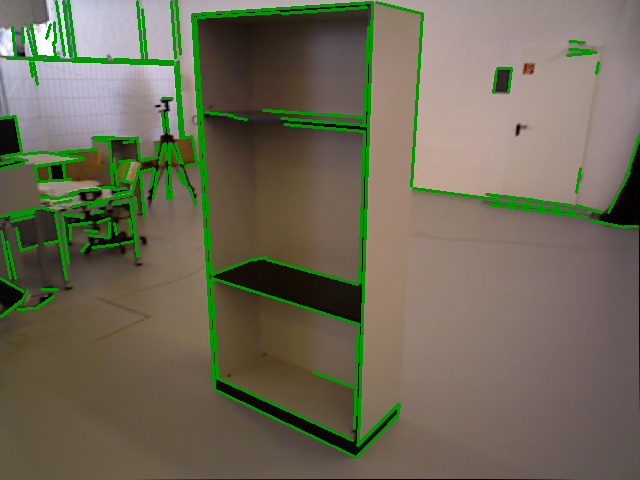}  }
  \subfigure[]
    { \includegraphics[width=45mm]{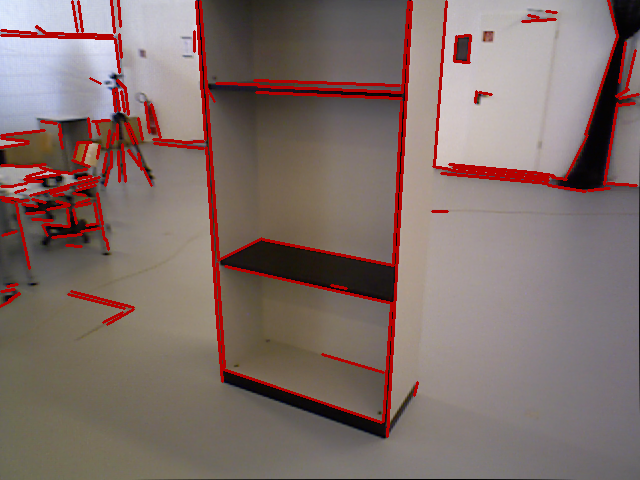}    } \\
  \subfigure[]
    {  \includegraphics[width=40mm]{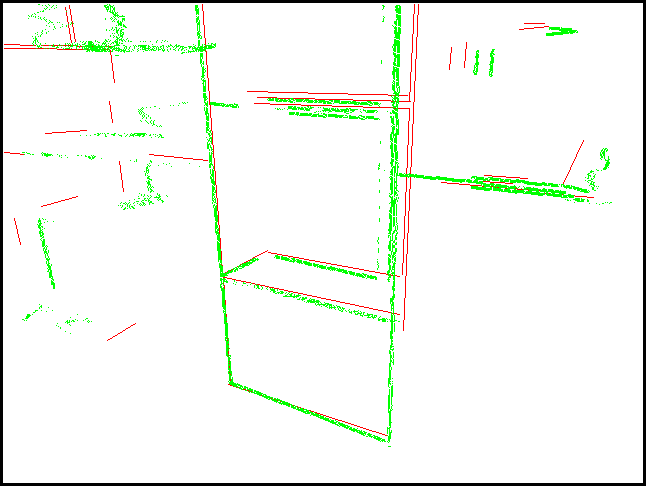}
       \includegraphics[width=40mm]{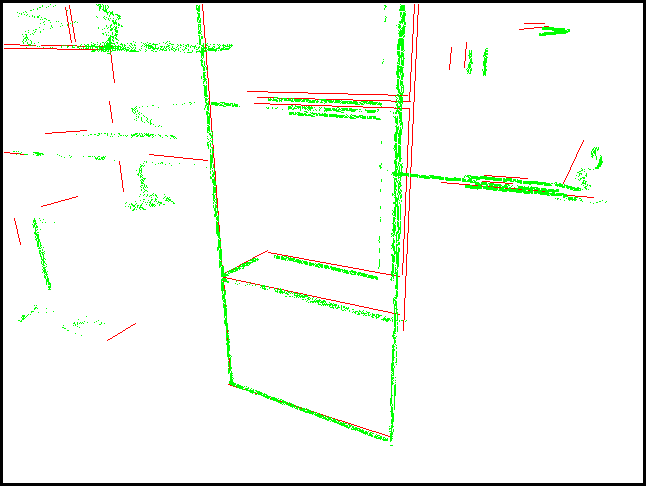}
       \includegraphics[width=40mm]{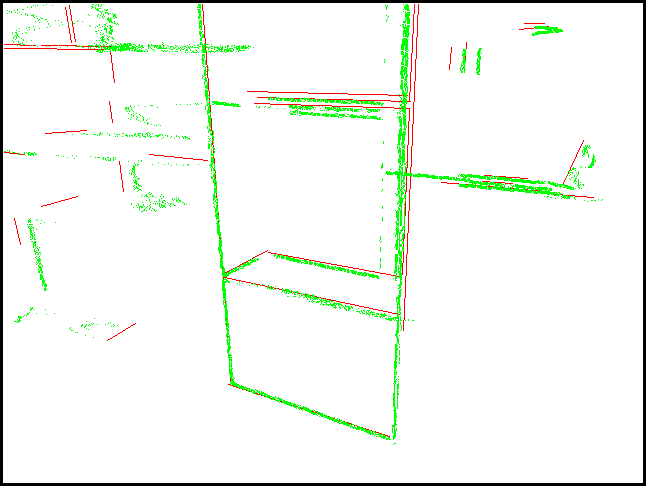}
       \includegraphics[width=40mm]{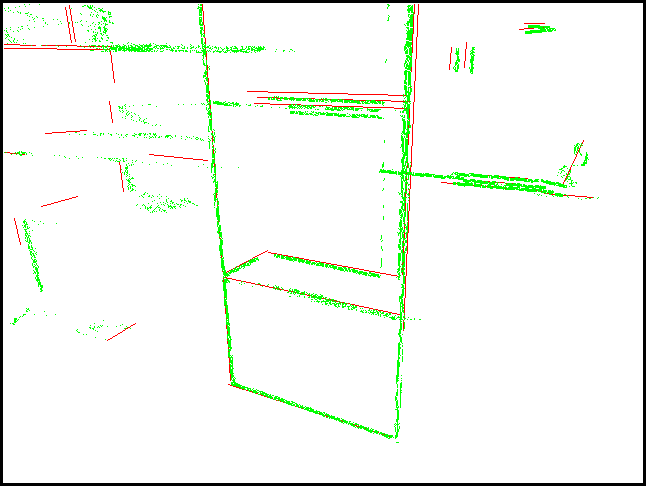}    }

\caption{Tracking iterations for two images in TUM \textit{fr3/cabinet\_big} dataset. (a) reference frame with detected edges (b) current frame with detected edges. Two frames are 41 frames apart (about 1.4s). (c) Re-projected pixels on current frame during optimization iterations corresponding to $1,4,9,20$. We can see that the re-projected edge pixels in green gradually align with the true edges in red. Best viewed in color.}
\label{fig:track iters}
\end{figure*}

\subsection{Notations}
We denote an intensity image as $I:\Omega \subset \mathbb{R}^2 \mapsto \mathbb{R}$, where $\Omega$ represents the image domain. We keep a per-pixel inverse depth map for a reference keyframe $D:\Omega \subset \mathbb{R}^2 \mapsto \mathbb{R^+}$ and inverse depth variance $V: \Omega \mapsto \mathbb{R}^+$.

The camera projection function is defined as $\pi: \mathbb{R}^3 \mapsto \mathbb{R}^2$, which projects a camera-centered 3D point onto image plane. The inverse projection function is then $\pi^{-1}: (\mathbb{R}^2,\mathbb{R}) \mapsto \mathbb{R}^3$ which back-projects an image pixel to the 3D space given its depth.

The transformation between the current frame and reference keyframe is defined by a rigid transformation $T \in SE(3)$. For an efficient optimization of $T$, we use the minimal manifold representation by elements of the Lie-algebra $\xi \in \mathfrak{se}(3)$ \cite{murray1994mathematical}, which is expressed by twist $\xi=(\mathbf{t};\mathbf{w})^T \in \mathbb{R}^6$. $\mathbf{t}\in \mathbb{R}^3$ is the translation component and $\mathbf{w} \in \mathbb{R}^3$ is the rotation component, which can form rotation matrix by the corresponding exponential map: $R(\mathbf{w})=\exp(\mathbf{[w]_\times}) \in SO(3)$.

A warping function is defined as $\tau: \Omega_1 \times \mathbb{R} \times \mathbb{R}^6 \mapsto \Omega_2$. It takes the parameters of a pixel $x \in \Omega_1$ in the first image $I_1$, its depth $d$ and relative camera transformation $\xi$, then returns the re-projection point in second image $I_2$. Internally, it first back-projects $x$ to 3D point by $\pi^{-1}$, transforms it using $\xi$, then projects to another frame using $\pi$.

An edge is represented by $L$ in 3D space and $l$ in 2D image plane. All the edge pixels in an image are defined as $M: \mathbb{R} ^2 \mapsto l$ which maps a pixel to its edge. Most of the edge pixels $M$ belong to the high gradient pixel set $\Omega$.

\section{Tracking}
\label{sec:tracking}
\subsection{Overview}
In the tracking thread, the depth map $D_{ref}$ of the reference frame $I_{ref}$ is assumed to be fixed. The current image $I$ is aligned by minimization of the photometric residual $r(\xi)$ and line re-projection geometric error $g(\xi)$ corresponding to two observation model: photometric intensity observation and edge position observations. It can be formulated as the following non-linear least squares problem:

\begin{equation}
\label{eq:all cost funcs}
E(\xi)= \sum_{i \in \Omega} r_i(\xi)^T \Sigma_{r_i}^{-1} r_i(\xi) + \sum_{j \in M} g_j(\xi)^T \Sigma_{g_j}^{-1} g_j(\xi)
\end{equation}
\noindent where photometric error $r_i$ is defined by \cite{engel2013semi} \cite{newcombe2011dtam}:
\begin{equation}
\label{eq:photo error}
r_i = I_{ref}(x_i) - I(\tau(x_i,D_{ref}(x_i),\xi))
\end{equation}

$g_j$ is the re-projection error of pixel $x_i$ to its corresponding line $l_j$ (homogeneous line representation):
\begin{equation}
\label{eq:line error}
g_j = l_j^T \ \widehat{\tau}(x_i,D_{ref}(x_i),\xi)
\end{equation}
where $\widehat{\tau}()$ is the homogeneous coordinate operation. This term is only used for the pixels of edges in $I_{ref}$ which also have a matching edge in $I$. $\Sigma_r$ and $\Sigma_g$ represents the uncertainty of two errors correspondingly.

The energy function Equation (\ref{eq:all cost funcs}) is minimized through iterative Gauss-Newton optimization. For iteration $n$, the small update is:
\begin{equation}
\delta \xi^{n} = - (J^TWJ)^{-1} J^T W \mathbf{E}(\xi^{n})
\end{equation}
\noindent Where $\mathbf{E}$ is the stacked error vectors composed of two parts: $\mathbf{E}=(r_1,...,r_n,g_1,...,g_m)^T$.
$J$ is the the Jacobian of $\mathbf{E}$ \textit{wrt}. $\xi$. $W$ is the weight matrix computed from uncertainty $\Sigma^{-1}$. A tracking illustration is shown in Fig \ref{fig:track iters}.

\subsection{Tracking uncertainty analysis}
\label{sec:line tracking uncertainty}
Combining different types of error terms in Equation (\ref{eq:all cost funcs}) increases the robustness and accuracy of pose estimation. The weights of different terms are proportional to the inverse of the error variance $\Sigma_r$ and $\Sigma_g$ computed from the observation models. Here, we provide an analysis of $\Sigma_g$. Photometric error uncertainty $\Sigma_r$ has been analysed in \cite{engel2013semi}.
 
In the general case, the uncertainty of the output of a function $f(x)$ propagated from the input uncertainty is expressed by:
\begin{equation}
\label{eq:error propa}
\Sigma_f \approx J_f \Sigma_x J_f^T
\end{equation}
\noindent where $J_f$ is the Jacobian of $f$ \textit{wrt.} $x$.

In our case, as defined in Equation (\ref{eq:line error}), the pixel re-projection error to line is a function of line equation $l_j$ and re-projected point $x_i'=\widehat{\tau}(x_i,D_{ref}(x_i),\xi)$. Line equation is computed by cross product of two line endpoints $ l_j = p_1 \times p_2$. We can assume that the uncertainties $\Sigma_p$ of end point positions $p_1$ and $p_2$ is bi-dimensional Gaussians with $\sigma=1$. Then we can use rule in Equation (\ref{eq:error propa}) to compute the uncertainties of line equation coefficients $l_j$. It basically implies that longer line has smaller line fitting uncertainties.

We can then similarly compute the variance of re-projection point $x_i'=\widehat{\tau}(x_i,D_{ref}(x_i),\xi)$. It is a function of pixel depth $D_{ref}(x_i)$ with variance $V_{ref}(x_i)$. The final re-projection error covariance is a combination of the two uncertainty sources: 
\begin{equation}
\Sigma_{g_j} = l_j^T \Sigma_{x_i'} l_j + x_i'^T \Sigma_{l_j} x_i'
\end{equation}

\vspace{0.2cm}


\section{Mapping}
\label{sec:mapping}
\subsection{Overview}
In the mapping thread, the depth map $D_{ref}$ of reference frame is updated through stereo triangulation in inverse depth filtering framework \cite{engel2013semi} followed by line regularization to improve the accuracy. The camera pose is assumed to be fixed in this step. The cost function for depth optimization is defined as follows:

\begin{equation}
\label{eq:map all cost}
E(D)= \sum_i r_i(d)^T \Sigma_{r_i}^{-1} r_i(d) + \sum_j G_j(d)^T \Sigma_{G_j}^{-1} G_j(d)
\end{equation}
\noindent where $r_i(d)$ is the stereo matching photometric error. SSD error over image patches is used to improve robustness. For a line $l_j$, we want its pixels to also form a line in 3D space after back-projection, so $G_j$ is edge regularization cost representing the distance of edge pixel's 3D point to 3D line. The regularization technique is also used in other dense mapping algorithms \cite{newcombe2011dtam} \cite{pinies2015dense}. If only the first term $r_i$ is used \cite{engel2013semi}, all the pixels are independent of each other and therefore could search independently along the epipolar line to find the matching pixel. Regularization term $G_j$ makes the depth of pixels on one edge correlate with each other and is typically solved by an iterative alternating optimization through duality principles \cite{newcombe2011dtam}. However, it requires much heavier computation. Instead, we optimize for $r_i$ and $G_j$ in two stages more efficiently.

\subsection{Stereo match with Lines}
\label{sec: line match}
For the pixels not on the edge or pixels on an edge which does not have a matching edge, we perform an exhaustive search for the stereo matching pixel by minimizing SSD error \cite{engel2013semi}. The depth interval for searching is limited by
$d+2\sigma_d$, where $d$ and $\sigma_d$ is the depth mean and standard deviation.

For the pixels with a matched edge, the re-projected points should lie on the matched edge as well as its epipolar line so we can directly compute their intersection as the matching point. We can also directly do line triangulation in Fig. \ref{fig:line triangulation} to compute all pixel's depth together. If the camera transform of current frame $I$ wrt. $I_{ref}$ is $R \in SO(3), \mathbf{t} \in \mathbb{R}^3$, then the 3D line $L$ can be represented the intersection of two back-projected plane \cite{hartley2003multiple}:

\begin{figure}[t]
\vspace{1.5 mm}
\centering
 \includegraphics[scale=0.4]{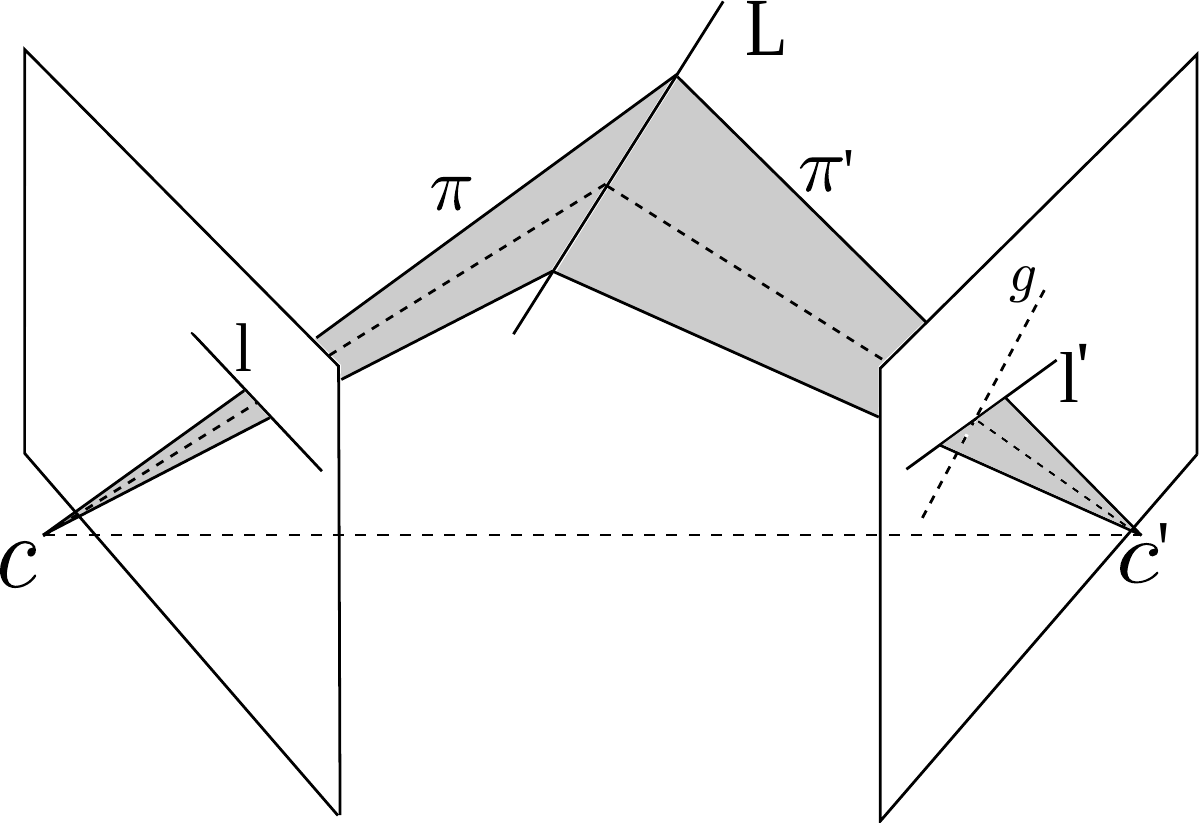}
\caption{Line triangulation. 3D line $L$ could be computed by the intersection of two back-projected planes $\pi$, $\pi'$. For each pixel on $l$, its stereo matching point is the intersection of epipolar line $g$ and matched edge $l'$. The triangulated point also lies on 3D line $L$. Modified from \cite{hartley2003multiple}.
}
\label{fig:line triangulation}
\end{figure}

\begin{equation}
L= \left[ \begin{matrix}
\pi_1^T \\
\pi_2^T
\end{matrix}  \right] = \left[  \begin{matrix}
l_1^TK   &0 \\
l_2^TKR   &l_2^T K \mathbf{t}
\end{matrix}  \right]
\end{equation}
\noindent where $l_1$ and $l_2$ are the line equation in $I_{ref}$ and $I$ respectively. $K$ is the intrinsic camera parameter. Then for each pixel, we can compute the intersection of the back-projected ray with $L$ to get its depth.

For the degenerated case where epipolar line and matched edge are (nearly) parallel, we cannot compute the 3D line accurately by plane intersection. Instead, we use the exhaustive search along the epipolar line to find the matching pixel with minimal SSD error.

\subsection{Line matching uncertainty analysis}
\label{sec: line match error}
\begin{figure}[t]
\vspace{1.5 mm}
\centering
 \includegraphics[scale=0.65]{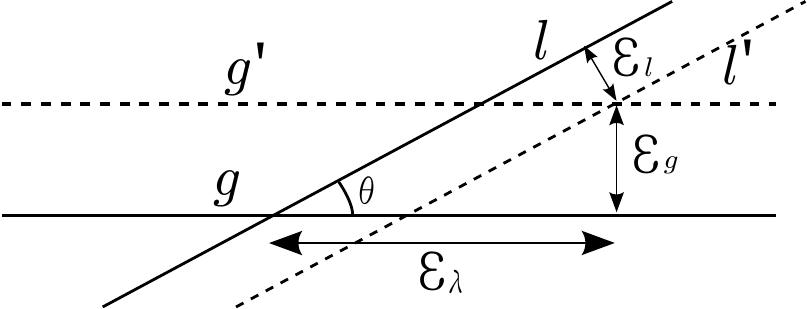}
\caption{Disparity error using line matching. $l$ is the edge where the matched pixel should lie. $g$ is the epipolar line. Due to a small positioning error $\epsilon_g$, $g$ is shifted to $g'$. The same with $l'$ of positioning error $\epsilon_l$.
The final resulting disparity error is $\epsilon_\lambda$.}
\label{fig:line uncertainty}
\end{figure}

The uncertainty of intensity based stereo searching along epipolar line has been analysed in \cite{engel2013semi}. Here we include the analysis of edge based stereo matching error. For each edge pixel in $I_{ref}$, denote its epipolar line in $I$ as $g$ and its matched edge as $l$ then the matching pixel is the intersection of $g$ and $l$. These two lines both have positioning error $\epsilon_l$ and $\epsilon_g$, and finally cause a disparity error $\epsilon_\lambda$ shown in Fig. \ref{fig:line uncertainty}. The edge uncertainty $\epsilon_l$ is already analyzed in Section \ref{sec:line tracking uncertainty} which is directly related to the edge length. $\epsilon_\lambda$ is large when $g$ and $l$ are nearly parallel. Mathematically we have:
\begin{equation}
\epsilon_\lambda = \epsilon_l / \sin(\theta)+ \epsilon_g \cot(\theta)
\end{equation}
where $\theta$ is the angle between line $l$ and $g$. 

From error propagation rule in Equation (\ref{eq:error propa}), we can compute the variance of the disparity error:
\begin{equation}
\sigma_{\lambda}^2 = \sigma_{l}^2 /\sin^2(\theta) + \sigma_{g}^2 \cot^2(\theta)
\end{equation}

Using the approximation that inverse depth $d$ is proportional to disparity $\lambda$, we can calculate the observation variance of $d$ using Equation (\ref{eq:error propa}). It can then be used to update the pixel's depth variance in a standard EKF filtering \cite{engel2013semi}.

\subsection{3D Line regularization}
Depth map regularization is important for monocular mapping approaches to improve the depth estimation accuracy. After the depth map EKF update in Section \ref{sec: line match error}, the pixels on a 2D edge may not correspond to a line in 3D space, therefore, we need to fit lines in 3D space and update a pixel's depth. 3D weighted line fitting is recently addressed in RGB-D line based odometry \cite{lu2015robust} which utilizes Levenberg-Marquardt iterative optimization to find the best 3D line. Here we propose a fast and analytical solution to the weighted 3D line fitting problem. 

Since the 3D points are back-projected from the same 2D edge, they should lie on the same plane $G$ from projective geometry. We can create anther coordinate frame $F$ whose $x,y$ axis lie on the plane $G$. The transformed point on the new coordinate frame is denoted as $p'$. We first use RANSAC to select a set of inlier 2D points. The metric for RANSAC is Mahalanobis distance, which is a weighted 
pixel to line Euclidean distance considering the uncertainty:

\begin{equation}
d_{\text{mah}}=\min_{q' \in l'} (p'-q')^T \Sigma_{p'}^{-1} (p'-q')
\end{equation}
where $q' \in l'$ indicates a point lying on line $l'$ in frame $F$. $d_{\text{mah}}$ could be computed analytically by taking the derivative \textit{wrt.} $q'$ and setting to zero. More details could be found in \cite{lu2015robust}.

After RANSAC, we can find the largest consensus set of points $p'_i$, $i=1,...,n$. This becomes a 2D weighted line fitting problem and we want to find the best line $L^*$ so that:
\begin{equation}
L^* = \min_{L} \sum_i \delta(p_i')^T \Sigma_{p_i'}^{-1} \delta(p_i')
\end{equation}
where $\delta(p_i')$ is distance of point $p_i'$ to line $L$ along $y$ axis. It is an approximation of point to line distance but could lead to a closed form solution. Stack all points $p'_i$ coordinates as $[\mathbf{X}, \mathbf{Y}]$ (after subtracting from mean) and weight matrix as $W$ which can be approximated as original image pixels' covariance. Then the line model under consideration is $\mathbf{Y}=\mathbf{X}\beta + \epsilon$, where $\beta$ is line coefficients, and $\epsilon$ is assumed to be normally distributed vector of noise. The MLE optimal line under Gaussian noise is:
\begin{equation}
\widehat{\beta} = arg\min_{\beta} \sum_i \epsilon_i^2 = (\mathbf{X}^T W \mathbf{X})^{-1} \mathbf{X}^T W \mathbf{Y}
\end{equation}

We can then transform the optimal line $L^*$ in coordinate frame $F$ back to the original camera optical frame and determine the pixel depth on the line.

\begin{figure}[t]
\vspace{1.5 mm}
\centering
 \includegraphics[scale=0.23]{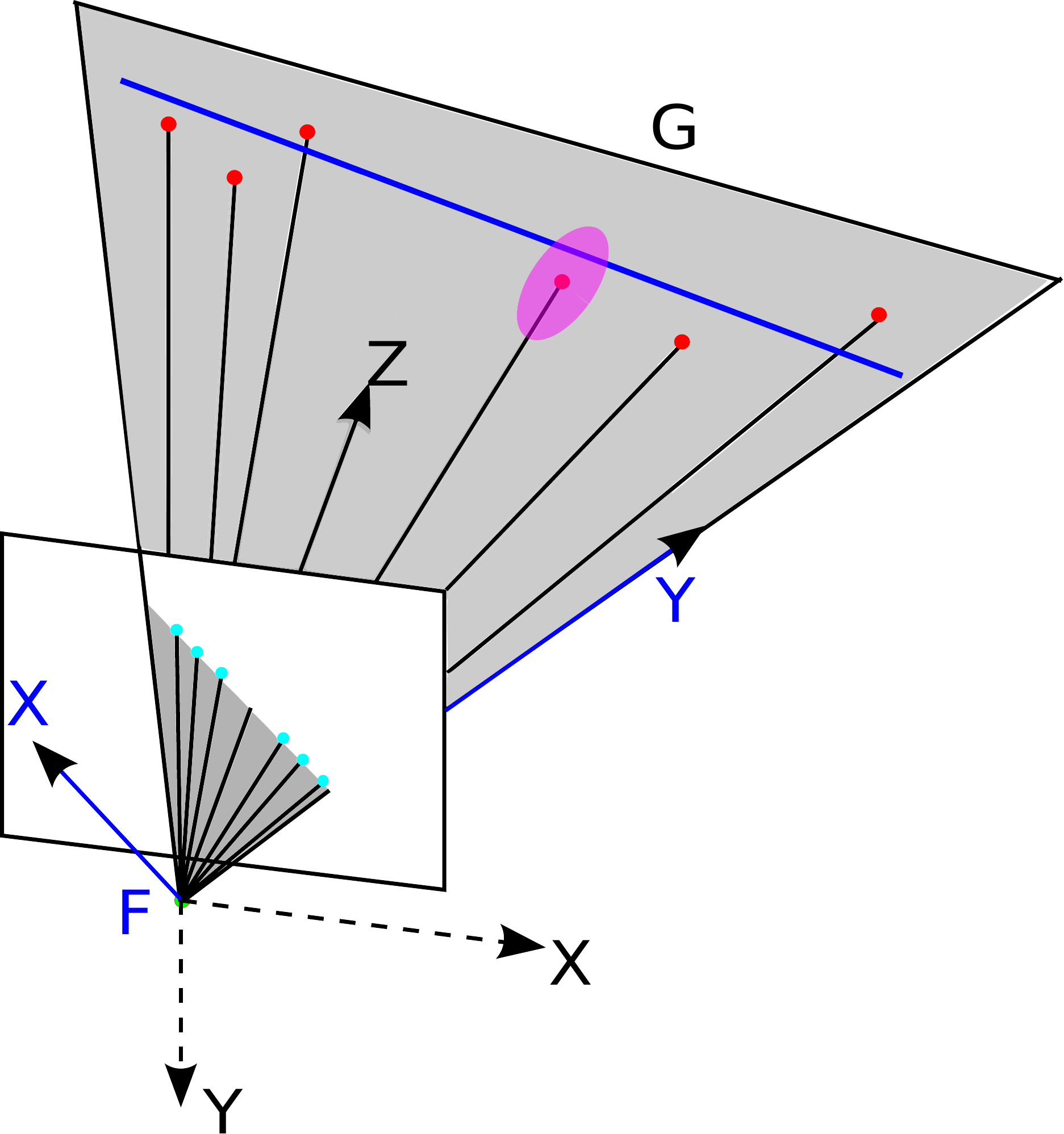}
\caption{3D line regularization. We first un-project pixel to 3D shown as red dots on the 3D plane $G$. The pink ellipse shows the uncertainty of 3D point. We can transform 3D points to a coordinate frame $F$ lying on the grey plane. The axis are $X$, $Y$ in blue. Then we can use RANSAC to analytically compute a weighted least square line instead of iterative optimization \cite{lu2015robust}. Image modified from \cite{lu2015robust}. Best viewed in color.}
\label{fig:initial_traj}
\end{figure}

\section{Experiments and results}
\label{sec:experiments}
\subsection{Implementation}
\label{sec:line implementation}
1. Edge detection and matching: We use the public line segment detection algorithm \cite{grompone2010lsd}. To improve the tracking accuracy, we adopt a coarse-to-fine approach using two pyramid levels with a scale factor of two. Due to the uncertainty of line detection algorithm, one complete line can sometimes break into multiple segments so we explicitly merge two lines whose angles and distance are very close within a threshold. After that, we need to remove very short line segments which may have large line fitting error. To speed up the line merging, lines are assigned to different bucket grids indexed by the middle points of an edge and the orientation of it. Then we only need to consider possible merging within the same and nearby bucket.

We then compute the LBD descriptor \cite{zhang2013efficient} for each line and match them across images. Bucket technique is also utilized to speed up the matching. Finally, line tracing is performed to find all pixels on an edge. We find that the system becomes more robust and accurate if we expand the line for one pixel possibly because more pixels are involved by the line constraints in tracking and mapping.


2. Keyframe-based VO: our approach doesn't have the bundle adjustment of points and lines in SLAM and SfM framework but could be extended to improve the performance. Camera tracking, line matching and stereo mapping are implemented only between the current frame and keyframe. 

\subsection{Experiments}
In this section, we test our algorithm on various public datasets including TUM RGBD \cite{sturm2012benchmark}, TUM mono \cite{engel2016monodataset} and ICL-NUIM \cite{handa:etal:ICRA2014}. We mainly compare with the state of art monocular direct SDVO \cite{engel2013semi} and feature based ORB SLAM \cite{mur2015orb}. We also provide some comparison with edge based VO \cite{jose2015realtime} \cite{gomez2016pl} in some datasets where the result is provided. For ORB SLAM, we turn off the loop closing thread, but still keep local and global bundle adjustment (BA) to detect incremental loop-closures while our algorithm and SDVO are VO algorithm without BA. We use the relative position error metric (RPE) by Strum \textit{et al} \cite{sturm2012benchmark}.

\begin{equation}
E_i = (Q_i^{-1}Q_{i+\delta})^{-1} (B_i^{-1}B_{i+\delta})
\end{equation}
\noindent where $Q_i \in SE(3)$ is the sequence of ground truth poses and $B_i \in SE(3)$ is the estimated pose. Scale is estimated to best align the trajectory.

\subsubsection{Qualitative results}
We choose TUM \textit{mono/38} \cite{engel2016monodataset} for VO and mapping visualization shown in Fig. \ref{fig:method overview}. It mainly contains homogeneous white surfaces but there are still many edges that could be utilized. Our method could generate good quality mapping and state estimation. More results can be found in the supplementary video.

\begin{figure}[t]
\vspace{1.5 mm}
\centering
\subfigure[]{
 \includegraphics[scale=0.15]{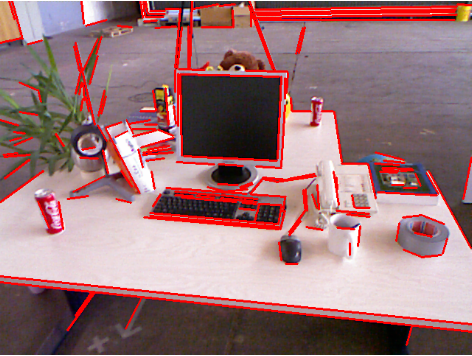}
 \label{fig:tum desk image}
 }
\subfigure[]{
 \includegraphics[scale=0.15]{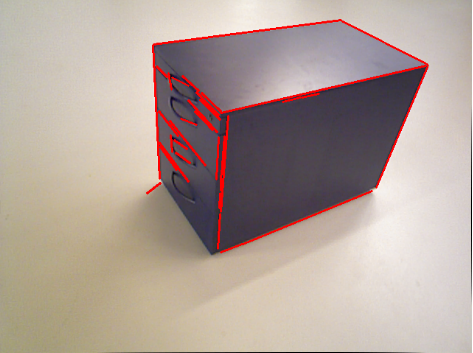}
 \label{fig:tum cabinet image}
 }
\subfigure[]{ 
 \includegraphics[scale=0.15]{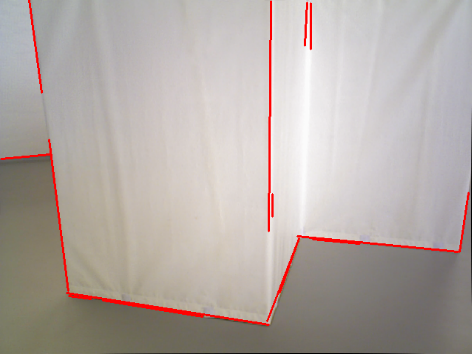}
 \label{fig:tum notex far} 
 }
\caption{Example images in TUM datasets with varying textures. (a) fr2/desk, (b) fr3/cabinet, (c) fr3/notex-far. ORB SLAM performs worse on (b) and (c) as there are fewer features points. Our algorithm can still utilize the matched edge features to improve the state estimation.}
\label{fig:tum sequence}
\end{figure}

\subsubsection{Quantitative results}

We first evaluate on two popular sequences of TUM RGBD dataset \textit{fr2/desk} and \textit{fr2/xyz} shown in Fig \ref{fig:tum desk image}. Comparison is shown in Table \ref{table:tum rgbd evaluation more comparison}, where result of SDVO and two edge based VO \cite{jose2015realtime} \cite{gomez2016pl} are obtained from their paper. These two scenarios are feature rich environments thus are most suitable for the feature-based ORB SLAM with BA. Due to the large amounts of high gradient pixels, SDVO also performs well. Due to many curved bottles, leafs, and small keyboards, there is relatively large line detection and matching errors for these environments, our algorithm performs similarly to SDVO but better than two other edge based VO.

\begin{table}[t]
\vspace{1.7 mm}
\caption{Relative Position Error (cm/s) Comparison on TUM Dataset}
\begin{center}
\begin{tabular}{c c c c c c}
\toprule
Sequence			& Ours		& SDVO  		& ORB-SLAM   	& \cite{jose2015realtime}   & \cite{gomez2016pl}	\\ \midrule
fr2/desk			& 1.88		& 2.1    	& \textbf{0.7}	& 2.8  		& 6.9 	\\ 
fr2/xyz         & 0.66		& 0.6    	& \textbf{0.6}	& 0.8  		& 2.1 	\\
\bottomrule
\end{tabular}
\end{center}
\label{table:tum rgbd evaluation more comparison}
\end{table}

We also provide results on more datasets shown in Table \ref{table:tum rgbd evaluation}, where other edge VO doesn't provide results. The top two scenes are relative easy environments. In TUM \textit{mono/38} in Fig. \ref{fig:method overview}, we only evaluate the beginning part which has ground truth pose. Since there are still some corner points on the door and showcase, ORB-SLAM with BA still performs the best but our algorithm clearly outperforms the SDVO. This is because the door surface is nearly homogeneous without large intensity gradients so the photometric error minimization of SDVO doesn't work very well while our algorithm can still use edges to minimize edge re-projection error.

The last four scenes in Table \ref{table:tum rgbd evaluation} are more challenging feature-less environments shown in Fig .\ref{fig:tum cabinet image} and Fig .\ref{fig:tum notex far}. ORB-SLAM doesn't work well and even fails (denoted as 'X') in some environments but direct VO can still work to some extent because direct methods utilize high gradient and edge pixels instead of feature points. Note that in \textit{ICL/office1} dataset, the overall scene has many feature points but ORB SLAM failure happens when the camera only observes white walls and ground with few distinguishable features. Our method with line clearly outperforms SDVO in most of the cases from the table and there are mainly two reasons. Firstly, by adding edges, we are utilizing more pixels for tracking. Some pixels might have low gradients due to homogeneous surfaces but can still be utilized because of lying on edges shown in Fig .\ref{fig:tum notex far}. Secondly, we are minimizing photometric error as well as geometric error, which is known to be more robust to image noise and has a large convergence basin. This has been analysed and verified in many other works \cite{engel2016monodataset} \cite{mur2015orb}.

\begin{table}[t]
\vspace{1.7 mm}
\caption{Relative Position Error (cm/s) Comparison on Various Datasets}
\begin{center}
\begin{tabular}{c c c c c c}
\toprule
Sequence			& Ours			& SDVO  			& ORB-SLAM   	\\ \midrule
ICL/office2     & 4.44			& 5.72    		& \textbf{2.11}	\\ \vspace{0.08cm}
mono/38		    & 2.04			& 5.4    		& \textbf{1.16}	\\ 
ICL/office1		& 1.85			& \textbf{1.33}	& X        		\\
fr3/cabinet-big & \textbf{8.82}	& 16.23  		& 33.57   		\\ 
fr3/cabinet     & \textbf{13.3}	& 21.7   		& X     			\\
fr3/notex-far   & \textbf{4.32}	& 10     		& X      		\\  
\bottomrule
\end{tabular}
\end{center}
\label{table:tum rgbd evaluation}
\end{table}

To demonstrate the advantage of a large convergence basin in the optimization, we select two frames from TUM \textit{fr3/cabinet\_big} which are 41 frames apart (1.3s) and show the tracking iterations in Fig. \ref{fig:track iters}. We can clearly see the re-projected pixels in green gradually align with the true edges in red.

\subsection{Time analysis}
We report the time usage of our algorithm running on TUM \textit{fr3/cabinet\_big} dataset shown in Table \ref{table:time analysis}. Using two octaves of line detection, there are totally 193 edges on average per frame. The total tracking thread apart from mapping takes 51.95 ms, able to run around 20Hz. Time could vary depending on the amounts of pixels involved in the optimization. For now, edge detection and descriptor computation consumes most of the time. This could be speeded up using down-sampled image. Edline edge detector \cite{akinlar2011edlines} can also be used to reduce detection time by half but it usually detects fewer edges compared to the currently used method \cite{grompone2010lsd} and may affect the state estimation accuracy in some challenging environments. Recently, Gomez \textit{et al} \cite{gomez2016pl} utilize edge tracking to decrease the computation instead of detection and matching for every frame, which is also a good solution.

\begin{table}[t]
\vspace{1.5 mm}
\caption{Time analyssi on TUM fr3/cabinet\_big dataset.}
\begin{center}
\begin{tabular}{c c c}
\toprule
Component	             &Value      \\ \midrule
Edge detection 		     & 16.24 ms   \\
Descriptor computation   & 11.95 ms   \\
Edge matching	      	& 4.52 ms    \\
Tracking	 time		    & 19.23 ms    \\
Mapping time			    & 10.99 ms    \\
Edge number             & 193     \\
\bottomrule
\end{tabular}
\end{center}
\label{table:time analysis}
\end{table}

\section{Conclusions}
\label{sec:conclusions}
In this paper, we propose a direct monocular odometry algorithm utilizing points and lines. We
follow the pipeline of SDVO \cite{engel2013semi} and add edges to improve both tracking and mapping 
performance. In the tracking part, we minimize both photometric error and geometric error
to the matched edges. In the mapping part, using matched edges, we can get stereo matching quickly and accurately without exhaustive search. An analytical solution is developed to regularize the depth map using edges.
We also provide probability uncertainty analysis of different observation models in tracking
and mapping part.

Our algorithm combines the advantage of direct and feature based VO. It is able to create a semi-dense map
and the state estimation is more robust and accurate due to the incorporation of edges and 
geometric error minimization. On various dataset evaluation, we achieve better or comparable performance than
SDVO and ORB SLAM. ORB SLAM with bundle adjustment works the best in environments with rich features. 
However, for scenarios with low texture, ORB SLAM might fail and direct methods usually work better. Our algorithm focuses 
on these scenarios and further improves the performance of SDVO by adding edges.

In the future, we want to reduce the computation of edge detection and matching by direct edge alignment. Also, 
bundle adjustment of edges in multiple frames could also be used to improve the accuracy. We will also exploit more 
information by combining points, edges, and planes \cite{syang2016popslam} in one framework to improve the accuracy and robustness in challenging environments.



\section*{ACKNOWLEDGMENTS}

This work is supported by NSF award IIS-1328930.


\bibliographystyle{unsrt}    
\bibliography{ref}

\addtolength{\textheight}{-12cm}   

\end{document}